\documentclass[11pt,a4paper]{article}
\usepackage[hyperref]{emnlp2020}
\usepackage{times}
\usepackage{latexsym}

\usepackage{algorithm} 
\usepackage{algorithmic}

\usepackage{tikz}
\usetikzlibrary{positioning,shapes,backgrounds}

\newcommand{\Images}{\mathcal{I}}
\newcommand{\Vocab}{\mathcal{V}}
\newcommand{\partition}{\mathbf{C}}

\usepackage{amssymb}
\usepackage{amsmath}

\let\emptyset\varnothing

\usepackage{graphicx}
\usepackage{booktabs}
\usepackage{float}

\usepackage{multirow}
\usepackage[normalem]{ulem}
\useunder{\uline}{\ul}{}

\usepackage{microtype}

\aclfinalcopy \def\aclpaperid{1597}

\newcommand{\OurTask}{ISIC}

\newcommand{\secref}[1]{Section~\ref{#1}}

\newcommand{\Figref}[1]{Figure~\ref{#1}}
\newcommand{\figref}[1]{Figure~\ref{#1}}

\newcommand{\Tabref}[1]{Table~\ref{#1}}
\newcommand{\tabref}[1]{Table~\ref{#1}}

\newcommand{\appref}[1]{Appendix~\ref{#1}}

\newcommand{\msg}{\mathbf{w}}
\newcommand{\state}{\mathbf{i}}
\newcommand{\given}{\mid}

\newcommand{\IssueSpeaker}{S_{1}^{\partition}}
\newcommand{\IssueEntropySpeaker}{S_{1}^{\partition + H}}

\newcommand{\TODO}[1]{#1} \newcommand{\EDIT}[1]{#1}

\title{Pragmatic Issue-Sensitive Image Captioning}

\author{  Allen Nie$^1$ \\\And 
  Reuben Cohn-Gordon$^2$ \\[1ex]
  $^1$Department of Computer Science ~~ $^2$Department of Linguistics \\
  Stanford University \\
  \texttt{\{anie, reubencg, cgpotts\}@stanford.edu} \\\And 
  Christopher Potts$^{1,2}$ \\
}  

\date{}

\begin{document}
\maketitle

\begin{abstract}
Image captioning systems need to produce texts that are not only \emph{true} but also \emph{relevant} in that they are properly aligned with the current issues. For instance, in a newspaper article about a sports event, a caption that not only identifies the player in a picture but also comments on their ethnicity could create  unwanted reader reactions. To address this, we propose Issue-Sensitive Image Captioning (\OurTask). In \OurTask, the captioner is given a target image and an \emph{issue}, which is a set of images partitioned in a way that specifies what information is relevant. For the sports article, we could construct a partition that places images into equivalence classes based on player position. To model this task, we use an extension of the Rational Speech Acts model. Our extension is built on top of state-of-the-art pretrained neural image captioners and explicitly uses image partitions to control caption generation. In both automatic and human evaluations, we show that these models generate captions that are descriptive and issue-sensitive. Finally, we show how \OurTask\ can complement and enrich the related task of Visual Question Answering.
\end{abstract}

\section{Introduction}\label{sec:intro}

Image captioning systems have improved dramatically over the last few years \citep{Karpathy:Li:2015,Vinyals-etal:2015,hendricks2016generating,rennie2017self,anderson2018bottom}, creating new opportunities to design systems that are not just accurate, but also produce descriptions that include relevant, characterizing aspects of their inputs. Many of these efforts are guided by the insight that high-quality captions are implicitly shaped by the communicative goal of identifying the target image up to some level of granularity \citep{vedantam2017context,mao2016generation,luo2018discriminability,cohn-gordon-etal-2018-pragmatically}.

\begin{figure}[!ht]
    \centering
    \includegraphics[width=0.95\linewidth]{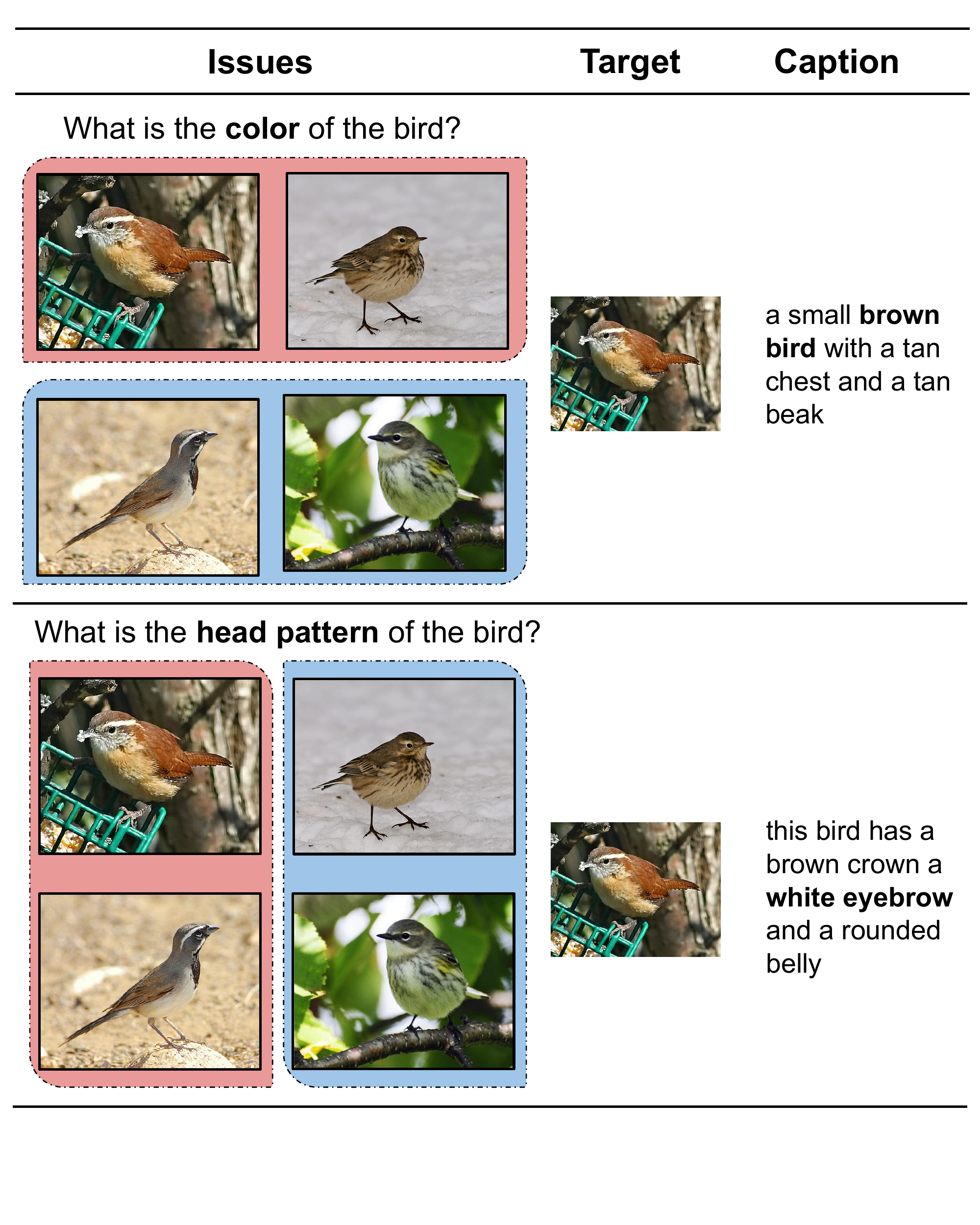}
    \caption{Examples highlighting the power of an issue-sensitive image captioner. Four images are partitioned in two ways, each capturing different issues by grouping them into equivalence classes. The first row contrasts the brown and grey color of the bird, and the second contrasts the existence of white eyebrows.  The target image is the same in both cases, but the partition leads to different captions that key into the structure of the input issue.}
    \label{fig:overview}
\end{figure}

In this paper, we seek to more tightly control the information that a pretrained captioner includes in its output texts. Our focus is on generating captions that are \emph{relevant} to the current issues. To see how important this can be, consider a newspaper article covering the action in a sports event. In this context, a caption that not only identified the player in a picture but also commented on their ethnicity could create unwanted reactions in readers, as it would convey to them that such information was somehow deemed relevant by the newspaper. On the other hand, in an article about diversity in athletics, that same caption might seem entirely appropriate.

To push captioners to produce more relevant texts, we propose the task of Issue-Sensitive Image Captioning (\OurTask). In \OurTask, the captioner's inputs are image/issue pairs, where an \emph{issue} is a set of images partitioned in a way that specifies what information is relevant. In our first example above, we might define a partition that grouped players into equivalence classes based on their team positions, abstracting away from other facts about them. For the second example, we might choose a more fine-grained partition based on position and demographic features. Given such inputs, the objective of the captioner is to produce a text that both accurately and uniquely describes the cell of the partition containing the target image. \Figref{fig:overview} illustrates with examples from our own models and experiments.

In defining the task this way, we are inspired by Visual Question Answering (VQA; \citealt{antol2015vqa}), but \OurTask\ differs from VQA in two crucial respects. First, we seek full image captions rather than direct answers. Second, our question inputs are not texts, but rather issues in the semantic sense: partitions on subsets of the available images. The \OurTask\ module reasons about the cells in these partitions as alternatives to the target image, and our notion of relevance is defined in these terms. Nonetheless, VQA and \OurTask\ complement each other: issues (as partitions) can be automatically derived from available image captioning and VQA datasets (\secref{sec:mscoco}), opening up new avenues for VQA as well.

Our models are built on top of pretrained image captioners with no need for additional training or fine-tuning. This is achieved by extending those models according to the Rational Speech Acts model (RSA; \citealt{Frank:Goodman:2012,Goodman:Stuhlmuller:2013}). RSA has been applied successfully to many NLP tasks (\secref{sec:related:rsa}). Our key modeling innovation lies in building issues into these models. In this, we are inspired by linguistic work on question-sensitive RSA \citep{Goodman:Lassiter:2013,Hawkins:Goodman:2019:Questions}.

Our central experiments are with the Caltech-UC San Diego-Bird dataset (CUB; \citealt{wah2011caltech}). This dataset contains extensive attribute annotations that allow us to study the effects of our models in precise ways. Using CUB, we provide quantitative evidence that our RSA-based models generate captions that both richly describe the target image and achieve the desired kinds of issue-sensitivity. We complement these automatic evaluation with a human evaluation in which participants judged our models to be significantly more issue-sensitive than standard image captioners. Finally, we show how to apply our methods to larger image captioning and VQA datasets that require more heuristic methods for defining issues. These experiments begin to suggest the potential value of issue-sensitivity in other domains that involve controllable text generation.
We share code for reproducibility and future development at \url{https://github.com/windweller/Pragmatic-ISIC}.

 \section{Related Work}\label{sec:related}

\subsection{Neural Image Captioning}

The task of image captioning crosses the usual boundary between computer vision and NLP; a good captioner needs to recognize coherent parts of the image and describe them in fluent text. \citet{Karpathy:Li:2015} and \citet{Vinyals-etal:2015} showed that large-capacity neural networks can get traction on this difficult problem. Much subsequent work has built on this insight, focusing on two aspects. The first is improving image feature quality \TODO{by using object-based features} \cite{anderson2018bottom}. The second is improving text generation quality by adopting techniques from reinforcement learning to directly optimize for the evaluation metric \cite{rennie2017self}. Our work rests on these innovations -- our base image captioning systems are those of \citet{hendricks2016generating} and \citet{rennie2017self}, which motivate and employ these central advancements.

\EDIT{There is existing work that proposes methods for controlling image caption generation with attributes. In general, these approaches involve models in which the attributes are part of the input, which requires a dataset with attributes collected beforehand. For instance,  \citet{mathews2015senticap} collected a small dataset with sentiment annotation for each caption, \citet{shuster2019engaging} collected captions with personality traits, and \citet{gan2017stylenet} with styles (such as humorous and romantic). The final metrics center around whether the human-generated caption was reproduced, or around other subjective ratings. By contrast, our method does not require an annotated dataset for training, and we measure the success of a model by whether it has resolved the issue under discussion.}

\subsection{Visual Question Answering}

In VQA, the model is given an image and a natural language question about that image, and the goal is to produce a natural language answer to the question that is true of the image \citep{antol2015vqa,goyal2017making}. This is a controllable form of (partial) image captioning. However, in its current form, VQA tends not to elicit linguistically complex texts; the majority of VQA answers are single words, and so VQA can often be cast as classification rather than sequence generation.
Our goal, in contrast, is to produce linguistically complex, highly descriptive captions. Our task additionally differs from VQA in that it produces a caption in response to an \emph{issue}, i.e., a partition of images, rather than a natural language question. 
In \secref{sec:S1} and \secref{sec:mscoco}, we describe how VQA and \OurTask\ can complement each other. 
\begin{figure*}[tp]
  \centering
  \clearpage{}\tikzset{
  every node=[draw, minimum height=4mm],
  redSquareLarge/.style={
    fill=red!40,
    minimum width=7mm,
    minimum height=7mm
  },    
  redSquareSmall/.style={
    fill=red!40,
    minimum width=4mm,
    minimum height=4mm
  },
  blueSquareLarge/.style={
    fill=blue!40,
    minimum width=7mm,
    minimum height=7mm
  },   
  blueSquareSmall/.style={
    fill=blue!40,
    minimum width=4mm,
    minimum height=4mm
  },
  greenCircleLarge/.style={
    fill=green!40,
    shape=circle,
    minimum width=7mm,
    minimum height=7mm
  },
  greenCircleSmall/.style={
    fill=green!40,
    shape=circle,
    minimum width=4mm,
    minimum height=4mm
  },
}

\begin{tikzpicture}
    
  \begin{scope}[local bounding box=group1]
    \node(p11){$\Bigg\{$};
    \node[right=-2mm of p11](p12){$\Big\{$};
    
    \node[redSquareSmall, right=0mm of p12](s11){};  
    \node[redSquareLarge, right=2mm of s11](s12){};
    
    \node[right=0mm of s12](p13){$\Big\}$};  
    \node[right=-1mm of p13](p14){$\Big\{$};
    
    \node[blueSquareSmall, right=0mm of p14](s13){};  
    \node[blueSquareLarge, right=2mm of s13](s14){};
    
    \node[right=0mm of s14](p15){$\Big\}$}; 
    \node[right=-1mm of p15](p16){$\Big\{$};
    
    \node[greenCircleSmall, right=0mm of p16](s15){};  
    \node[greenCircleLarge, right=2mm of s15](s16){};
    
    \node[right=0mm of s16](p17){$\Big\}$};
    \node[right=-2mm of p17](p18){$\Bigg\}$};
  \end{scope}

  \node[redSquareSmall, right=10mm of p18](t1){};

  \node[right=10mm of t1](caption1){``A red square''};
  
      \node[below=1mm of p11](p21){$\Bigg\{$};
  \node[right=-2mm of p21](p22){$\Big\{$};
  
  \node[redSquareSmall, right=0mm of p22](s21){};  
  \node[blueSquareSmall, right=2mm of s21](s22){};
  
  \node[right=0mm of s22](p23){$\Big\}$};  
  \node[right=-1mm of p23](p24){$\Big\{$};
  
  \node[redSquareLarge, right=0mm of p24](s23){};  
  \node[blueSquareLarge, right=2mm of s23](s24){};
  
  \node[right=0mm of s24](p25){$\Big\}$}; 
  \node[right=-1mm of p25](p26){$\Big\{$};
  
  \node[greenCircleSmall, right=0mm of p26](s25){};  
  \node[greenCircleLarge, right=2mm of s25](s26){};
  
  \node[right=0mm of s26](p27){$\Big\}$};
  \node[right=-2mm of p27](p28){$\Bigg\}$};

  \node[redSquareSmall, right=10mm of p28](t2){};

  \node[right=10mm of t2](caption2){``A small square''};

  \node[above=4mm of t1](Target){\textbf{Target}};
  \node[] at (Target -| caption1){\textbf{Caption}};
  \node[] at (Target -| group1){\textbf{Issue}};
\end{tikzpicture}\clearpage{}
  
  \vspace{-5mm}   
  \caption{Two idealized examples highlighting the desired behavior for \OurTask. A single set of images is partitioned in two ways. The top row groups them by color and shape, whereas the bottom row groups them by size and shape. A successful system for \OurTask\ should key into these differences: for the same target image, its captions should reflect the partition structure and identify which cell the target belongs to, as in our examples. A caption like ``A square'' would be inferior in both contexts because it doesn't convey which cell the target image belongs to.}
  \label{fig:toy-example}
\end{figure*}
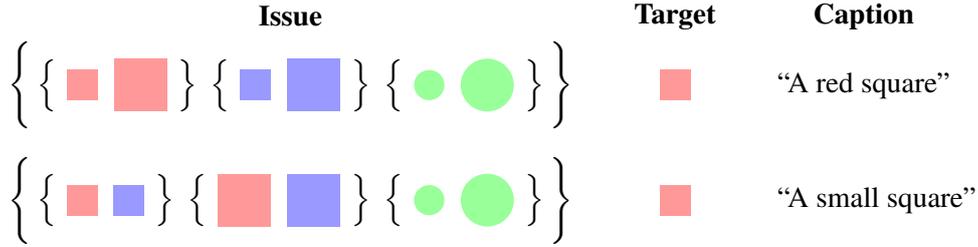

\subsection{The Rational Speech Acts Model}\label{sec:related:rsa}

The Rational Speech Acts model (RSA) was developed by \citet{Frank:Goodman:2012} with important precedents from \citet{Lewis69}, \citet{Jaeger:2007}, \citet{Franke09DISS}, and \citet{golland-etal-2010-game}. RSA defines nested probabilistic speaker and listener agents that reason about each other in communication to enrich the basic semantics of their language. The model has been applied to a wide variety of diverse linguistic phenomena. Since RSA is a probabilistic model of communication, it is amenable for incorporation into many modern NLP architectures. A growing body of literature shows that adding RSA components to NLP architectures can help them to capture important aspects of context dependence in language, including referential description generation \citep{Monroe:Potts:2015,andreas-klein-2016-reasoning,monroe-etal-2017-colors}, instruction following \citep{fried-etal-2018-unified}, collaborative problem solving \citep{Tellex-etal:2014}, and translation \citep{cohn-gordon-goodman-2019-lost}. 

Broadly speaking, there are two kinds of approaches to incorporating RSA into NLP systems. One class  performs end-to-end learning of the RSA agents \citep{Monroe:Potts:2015,mao2016generation,White-etal:2020}. The other uses a pretrained system and applies RSA at the decoding stage \citep{andreas-klein-2016-reasoning,vedantam2017context,monroe-etal-2017-colors,fried-etal-2018-unified}. We adopt this second approach, as it highlights the ways in which one can imbue a wide range of existing systems with new capabilities.

\subsection{Issue-Sensitivity in Language}\label{sec:related:issues}

Our extension of RSA centers on what we call \emph{issues}. In this, we build on a long tradition of linguistic research on the ways in which language use is shaped by the issues (often called Questions Under Discussion) that the discourse participants regard as relevant \citep{Groenendijk84,Ginzburg96DYNAMICS,Roberts96}.  Issues in this sense can be reconstructed in many ways. We follow \citet{Lewis88RELEVANT} and many others in casting an issue as a partition on a space of states into cells. 
Each cell represents a possible resolution of the issue. These ideas are brought into RSA by \citet{Goodman:Lassiter:2013} and \citet{Hawkins:Goodman:2019:Questions}. We translate those ideas into the models for \OurTask\ (\secref{sec:models}), where an issue takes the form of a partition over a set of natural images.

 \section{Task Formulation}\label{sec:task}

In standard image captioning, the input $\state$ is an image drawn from a set of images $\Images$, and the output $\msg$ is a sequence of tokens $[w_{1}, \ldots, w_{n}]$ such that each $w_{i} \in \Vocab$, where $\Vocab$ is the vocabulary.

In \OurTask, we extend standard image captioning by redefining the inputs as pairs $(\partition, \state)$, where $\partition$ is a partition\footnote{A set of sets $\mathbf{X}$ is a partition of a set $X$ iff $u \cap v = \emptyset$ for all $u, v \in \mathbf{X}$ and $\bigcup_{u \in \mathbf{X}} = X$.} on a subset of elements of $\Images$ and $\mathbf{i} \in \bigcup_{u\in \partition}$. We refer to the partitions $\partition$ as \emph{issues}, for the reasons discussed in \secref{sec:related:issues}. The goal of \OurTask\ is as follows: given input $(\partition, \state)$, produce a caption $\msg$ that provides a true resolution of $\partition$ for $\state$, which reduces to $\msg$ identifying the cell of $\partition$ that contains $\state$, as discussed in \secref{sec:related:issues}. \Figref{fig:toy-example} presents an idealized example. (\Figref{fig:overview} is a real example involving CUB and an \OurTask\ captioner; see also \figref{fig:cub-example} and \figref{fig:mscoco-examples} below.)

In principle, we could try to learn this kind of issue sensitivity directly from a dataset of examples $((\partition, \state), \msg)$. We do think such dataset could be collected, as discussed briefly in \secref{sec:conclusion}. However, such datasets would be very large (each image needs to collect $|\partition|$ number of captions), and our primary modeling goal is to show that such datasets need not be created. The issue-sensitive pragmatic model we introduce next can realize the goal of \OurTask\ without training data of this kind.

\section{Models}\label{sec:models}

\subsection{Neural Pragmatic Agents}\label{sec:prag}

The models we employ for \OurTask\ define a hierarchy of increasingly sophisticated speaker and listener agents, in ways that mirror ideas from Gricean pragmatics \citep{Grice75} about how meaning can arise when agents reason about each other in both production and comprehension (see also \citealt{Lewis69}).

Our base agent is a speaker $S_{0}(\msg\given\state)$. In linguistic and psychological models, this agent is often defined by a hand-built semantics.
In contrast, our $S_{0}$ is a trained neural image captioning system. As such, they are learned from data, with no need to hand-specify a semantic grammar or the like.

The pragmatic listener $L_{1}(\state\given\msg)$ defines a distribution over states $\state$ given a message $\msg$. The distribution is defined by applying Bayes' rule to the $S_{0}$ agent:
\begin{equation}\label{eq:L1}
  L_{1}(\state\given\msg) =
  \frac{
    S_{0}(\msg\given\state)P(\state)
  }{
    \sum_{\state' \in \Images} S_{0}(\msg\given\state')P(\state')
  }
\end{equation}
where $P(\state)$ is a prior over states $\state$ (always flat in our work). This agent is pragmatic in the sense that it reasons about another agent, showing behaviors that align with the Gricean notion of \emph{conversational implicature} \citep{Goodman:Frank:2016}.

We can then define a pragmatic speaker using a utility function $U_1$, in turn defined in terms of $L_{1}$:
\begin{align}
\label{eq:U1}
U_{1} (\state,\msg)
&= \log L_1(\state\given\msg)
\\
\label{eq:S1-simple}
S_1(\msg \given \state)  
&= 
\frac{
    \exp\left(\alpha U_1(\state,\msg)-\mathit{cost}(\msg)\right)
}{
    \sum_{\msg'} 
    \exp\left(\alpha U_1(\state,\msg')-\mathit{cost}(\msg')\right)
}
\end{align}
Here, $\alpha$ is a parameter defining how heavily $S_{1}$ is influenced by $L_{1}$. The term  $\mathit{cost}(\msg)$ is a cost function on messages. In other work, this is often specified by hand to capture analysts' intuitions about complexity or markedness. In contrast, our version is entirely data-driven: we specify $\mathit{cost}(\msg)$ as ${-}\log(S_0(\msg\given\state))$.

\subsection{Issue-Sensitive Speaker Agents}\label{sec:S1}

The agent in \eqref{eq:S1-simple} has been widely explored and shown to deliver a powerful notion of context dependence \citep{andreas-klein-2016-reasoning,monroe-etal-2017-colors}. However, it is insensitive to the issues $\partition$ that characterize \OurTask. To make this connection, we extend \eqref{eq:S1-simple} with a term for these issues:
\begin{align}
U_1^{\partition}(\state,\msg,\partition) &= \log\left(\sum_{\state' \in \Images}
  \delta_{[\partition(\state)= \partition(\state')]}
  L_1(\state'\given\msg)
  \right)\label{eq:U1C}\\
\IssueSpeaker(\msg \given \state,\partition)  
&\propto \exp\left(\alpha  U_1^{\partition}(\state,\msg,\partition)-\mathit{cost}(\msg)\right)
\label{eq:S1}
\end{align}
where $\delta_{[\partition(\state) = \partition(\state')]}$ is a partition function, returning $1$ if $\state$ and $\state'$ are in the same cell in $\partition$, else $0$. This is based on a similar model of \citet{kao2014nonliteral}. We use $\partition(\state)$ to denote the cell to which image $\state$ belongs under $\partition$ (a slight abuse of notation, since $\partition$ is a set of sets).

The construction of the partitions $\partition$ is deliberately left open at this point. In some settings, the set of images $\Images$ will have metadata that allows us to construct these directly. For example, in the CUB dataset, we can use the attributes to define intuitive partitions directly -- e.g., the partition that groups images into equivalence classes based on the beak color of the birds they contain. The function can also be parameterized by a full VQA model $\mathcal{A}$. For a given question text $\mathbf{q}$ and image $\mathbf{i}$, $\mathcal{A}$ defines a map from $(\mathbf{q}, \mathbf{i})$ to answers $\mathbf{a}$, and so we can partition a subset of $\Images$ based on equivalence classes defined by these answers $\mathbf{a}$.

\subsection{Penalizing Misleading Captions}\label{sec:entropy}

The agent in \eqref{eq:S1} is issue-sensitive in that it favors messages that resolve the issue $\partition$. However, it does not include a pressure against hyper-specificity; rather, it just encodes the goal of identifying partition cells. This poses two potential problems.

The first can be illustrated using the top row of \figref{fig:toy-example}. All else being equal, our agent \eqref{eq:S1} would treat ```A red square'' and ``A small red square'' as equally good captions, even though the second includes information that is intuitively gratuitous given the issue. This might seem innocent here, but it can raise concerns in real environments,
as we discussed in \secref{sec:intro} in connection with our newspaper article examples.

The second problem relates to the data-driven nature of the systems we are developing: in being hyper-specific, we observed that they often mentioned properties not true of the target but rather only true of members of their equivalence classes. For example, in \figref{fig:toy-example}, the target could get incorrectly described with ``A large red square'' because of the other member of its cell.

We propose to address both these issues with a second utility term $U_{2}$:
\begin{equation}
    \label{eq:H}
    U_2(\msg, \state, \partition) = 
    H(L_1(\state' \given \msg)
    \cdot 
    \delta_{[\partition(\state)= \partition(\state')]})
\end{equation}
where $H$ is the information-theoretic entropy. This encodes a pressure to choose utterances which result in the $L_0$ spreading probability mass as evenly as possible over the images in the target image cell. This discourages very specific descriptions of any particular image in the target cell, thereby solving both of the problems we identified above.

We refer to this agent as $\IssueEntropySpeaker$. Its full specification is as follows:
\begin{multline}\label{eq:S1H}
\IssueEntropySpeaker(\msg \given \state,\partition)  \propto\\ 
\exp\left(\alpha \left( (1 - \beta) U_1 + \beta U_{2}\right)-\mathit{cost}(\msg)\right)
\end{multline}
where $\beta \in [0,1]$ is a hyperparameter that allows us to weight these two utilities differently.

\subsection{Reasoning about Alternative Captions}\label{sec:alternatives}

A pressing issue which arises when computing probabilities using \eqref{eq:S1-simple}, \eqref{eq:S1}, and \eqref{eq:S1H} is that the normalization constant includes a sum over all possible captions $\msg'$. In the present setting, the set of possible captions is infinite (or at least exponentially large in the maximum caption length), making this computation intractable.

There are two solutions to this intractability proposed in the literature: one is to use $S_{0}$ to sample a small subset of captions from the full space, which then remains fixed throughout the computation \citep{andreas-klein-2016-reasoning,monroe-etal-2017-colors}. The drawback of this approach is that the diversity of captions that the $S_1$ can produce is restricted by the $S_0$. Since our goal is to generate captions which may vary considerably depending on the issue, this is a serious limitation.

The other approach is to alter the model so that the RSA reasoning takes place greedily during the generation of each successive word, word piece, or letter in the caption, so that the possible ``utterances'' at each step are drawn from a relatively small set of options to avoid exponential increase in search space \citep{cohn-gordon-etal-2018-pragmatically}. We opt for this \emph{incremental} formulation and provide the full details on this model in \appref{app:incremental}.

 \section{CUB Experiments}\label{sec:cub}

\subsection{Preliminaries}\label{sec:cub:prelims}

\paragraph{Dataset}

The Caltech UC San Diego-Bird (CUB) dataset contains 11,788 images for 200 species of North American birds \cite{wah2011caltech}. Each image contains a single bird and is annotated with fine-grained information about the visual appearance of that bird, using a system of 312 attributes (all of them binary) devised by ornithologists. The attributes have a \emph{property::value} structure, as in \textit{has\_wing\_color::brown}, and are arranged hierarchically from high-level descriptors (e.g., \emph{bill}) to very specific low-level attributes (e.g., \emph{belly pattern}). Appendix~\ref{app:cub} provides a detailed example.

\citet{reed2016learning} annotated each image in CUB with five captions. These captions were generated by crowdworkers who did not have access to the attribute annotations, and thus they vary widely in their alignment with the CUB annotations.

\paragraph{Constructing CUB Partitions}

CUB is ideal for testing our issue-sensitive captioning method because we can produce partitions directly from the attributes. For example,
\textit{has\_wing\_color::brown} induces a binary partition
into birds with brown wings and birds with non-brown wings, and \textit{has\_wing\_color} alone induces a partition that groups birds into equivalence classes based on their wing-color values. We selected the 17 most frequently appearing attributes, which creates 17 equivalence classes to serve as our issues.

\begin{figure*}[tp]
\centering
\includegraphics[width=0.95\linewidth]{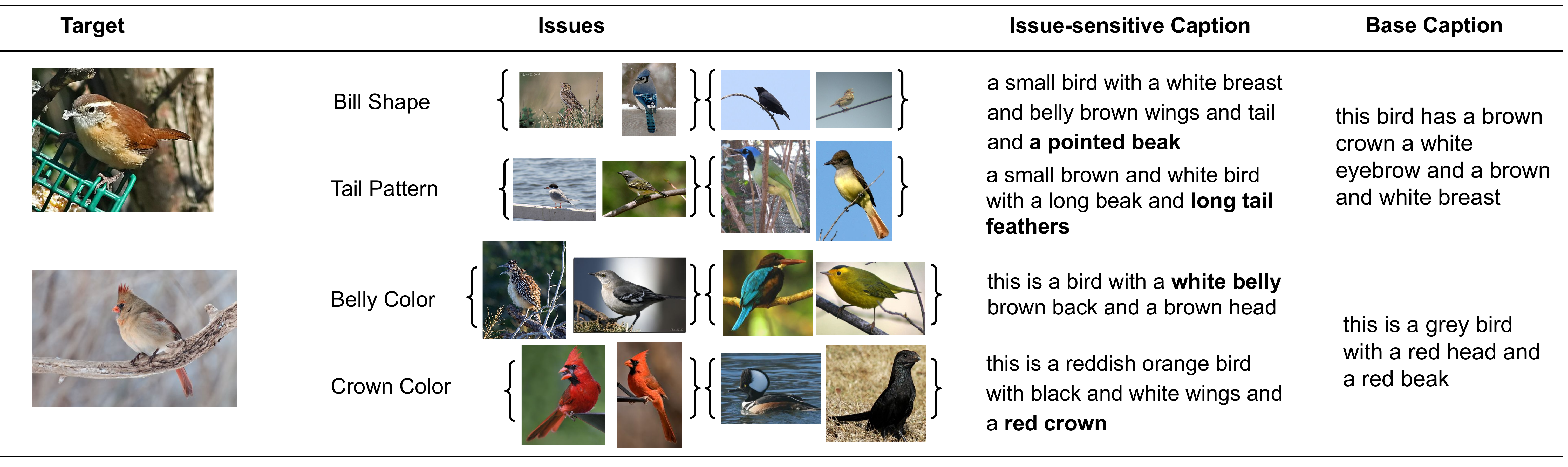}
    \caption{Captions for CUB. The left-hand cell in the `Issues' column contains the target image, and right-hand cell is the union of the distractor cells. The `Base Caption' texts are those produced by our $S_{0}$ model, and the `Issue-sensitive Caption' texts were produced by $\IssueEntropySpeaker$.}
    \label{fig:cub-example}
\end{figure*}

\paragraph{Base Captioning System}

We trained a model released by \citet{hendricks2016generating} with the same data-split scheme, where we have 4,000 images for training, 1,994 images for validation, and 5,794 images for testing. The model is a two-layer long short-term memory model \citep{Hochreiter:Schmidhuber:1997} with 1000-dimensional hidden size and 1000-dimensional word embeddings. We trained for 50 epochs with a batch size of 128 and learning rate 1$e{-}$3. The final CIDEr score for our model is 0.52 on the test split. We use greedy decoding to generate our captions.

\paragraph{Feature-in-Text Classifier}
In order to examine the effectiveness of our issue-sensitive captioning models, we need to be able to identify whether the generated caption contains information regarding the issue. Even though each CUB image has a complete list of features for its bird, we must map these features to descriptions in informal text. 
For this, we require a text classifier. Unfortunately, it is not possible to train an effective classifier on the CUB dataset itself. As we noted above, the caption authors did not have access to the CUB attribute values, and so their captions tend to mention very different information than is encoded in those attributes. Furthermore, even if we did collect entirely new captions with proper attribute alignment, the extreme label imbalances in the data would remain a challenge for learning.

To remedy this, we use a sliding window text classifier. First, we identify keywords that can describe body parts (e.g. ``head'', ``malar'', ``cheekpatch'') and extract their positions in the text. Second, we look for keywords related to aspects (e.g., ``striped'', ``speckled''); if these occur before a body-part word, we infer that they modify the body part. Thus, for example, if ``scarlet and pink head'' is in the caption, then we infer that it resolves an issue about the color of the bird's head.

This classifier is an important assessment tool for us, so it needs to be independently validated. We meet this need using our human study in \secref{sec:human:studies}, which shows that our classifier is extremely accurate and, more importantly, not biased towards our issue-sensitive models.

\subsection{Evaluating Attribute Coverage}

We begin by assessing the extent to which our issue-sensitive pragmatic models produce captions that are more richly descriptive of the target image than a base neural captioner $S_{0}$ and its simple pragmatic variant $S_{1}$. For CUB, we can simply count how many attributes the caption specifies according to our feature-in-text classifier. More precisely, for each image and each model, we generate captions under all resolvable issues, concatenate those captions, and then use the feature-in-text classifier to obtain a list of attributes, which we can then compare to the ground truth for the image as given by the CUB dataset. 

\EDIT{For $S_{0}$ and $S_{1}$, the captions do not vary by issue, whereas our expectation is that they do vary for $\IssueSpeaker$ and $\IssueEntropySpeaker$. To further contextualize the performance of the issue-sensitive agents, we additionally define a model $S_{0}$ Avg that takes as inputs the average of all the features from all the images in the current partition, and otherwise works just like $S_{0}$. This introduces a rough form of issue-sensitivity, allowing us to quantify the value of the more refined approach defined by $\IssueSpeaker$ and $\IssueEntropySpeaker$. Appendix~\ref{app:optimization} provides full details on how these models were optimized.}

\Tabref{tab:attribute-coverage} reports on this evaluation. Precision for all models is very high; the underlying attributes in CUB are very comprehensive, so all high-quality captioners are likely to do well by this metric. In contrast, the recall scores vary substantially, and they clearly favor the issue-sensitive models, revealing them to be substantially more descriptive than $S_{0}$ and $S_{1}$. \Figref{fig:cub-example} provides examples that highlight these contrasts: whereas the $S_{0}$ caption is descriptive, it simply doesn't include a number of attributes that we can successfully coax out of an issue-sensitive model by varying the issue. 

\EDIT{The results also show that $\IssueSpeaker$ and $\IssueEntropySpeaker$ provide value beyond simply averaging image features in $\partition$, as they both outperform $S_0$~Avg. However, it is noteworthy that even the rough notion of issue-sensitivity embodied by $S_0$~Avg seems beneficial.}

\EDIT{\Tabref{tab:body-part-aspect-per-image} summarizes attribute coverage at the level of individual categories, for our four primary models.  We see that the issue-sensitive models are clear winners. However, the entropy term in $\IssueEntropySpeaker$ seems to help for some categories but not others, suggesting underlying variation in the categories themselves.}

\begin{table}[tp]
\centering
\begin{tabular}{l c c c}
\toprule
   &  Precision & Recall & F1 \\
\midrule
$S_0$                  & \textbf{96.1} & 16.6 & 28.3  \\
$S_0$ Avg & \EDIT{94.0} & \EDIT{32.7} & \EDIT{48.5} \\
$S_1$                  & 93.9 & 29.6 & 45.0  \\
$\IssueSpeaker$        & 94.8 & 44.3 & 60.4 \\
$\IssueEntropySpeaker$ & 94.7 & \textbf{50.8} & \textbf{66.2} \\ 
\bottomrule
\end{tabular}
\caption{Attribute coverage results.} \label{tab:attribute-coverage}
\end{table}

\subsection{Evaluating Issue Alignment}\label{sec:cub:issues}

Our previous evaluation shows that varying the issue has a positive effect on the captions generated by our issue-sensitive models, but it does not assess whether these captions resolve individual issues in an intuitive way. We now report on an assessment that quantifies issue-sensitivity in this sense.

The question posed by this method is as follows: for a given issue $\partition$, does the produced caption precisely resolve $\partition$? We can divide this into two sub-questions. First, does the caption resolve $\partition$, which is a notion of \emph{recall}. Second, does the caption avoid addressing issues that are distinct from $\partition$, which is a notion of \emph{precision}. The recall pressure is arguably more important, but the precision one can be seen as assessing how often the caption avoids irrelevant and potentially distracting information, as discussed in \secref{sec:entropy}.

\Tabref{tab:resolution-results} reports on this issue-sensitive evaluation, with $F_{1}$ giving the usual harmonic mean between our versions of precision and recall. Overall, the scores reveal that this is a very challenging problem, which traces to the fine-grained issues that CUB supports. Our $\IssueEntropySpeaker$ agent is nonetheless definitively the best, especially for recall.

\begin{table}[tp]
\centering
\begin{tabular}{l c c c c}
\toprule
Issues & $S_0$ & $S_1$ & $\IssueSpeaker$ & $\IssueEntropySpeaker$ \\
\midrule
wing pattern     & 3.9   & 22.1  & \textbf{28.7}    & 14.9        \\
belly pattern    & 7.6   & 19.4  & 32.0    & \textbf{39.8}        \\
breast pattern   & 7.6   & 17.0  & 30.8    & \textbf{35.2}        \\
nape color       & 6.0   & 18.1  & 31.2    & \textbf{49.2}        \\
upper tail color & 1.3   & 24.5  & \textbf{30.6}    & 22.6        \\
under tail color & 1.5   & 26.3  & \textbf{33.1}    & 22.9        \\
back color       & 5.1   & 24.9  & 32.6    & \textbf{66.7}        \\
leg color        & 6.2   & \textbf{51.0}  & 45.6    & 7.4         \\
throat color     & 16.4  & 46.2  & 66.0    & \textbf{69.8}        \\
crown color      & 49.5  & 50.4  & 77.4    & \textbf{90.4}        \\
bill shape       & 47.7  & 43.7  & 71.1    & \textbf{91.8}        \\
eye color        & 24.4  & 54.1  & \textbf{61.3}    & 53.3        \\
wing color       & 39.2  & 70.5  & \textbf{82.7}    & 77.4        \\
bill color       & 38.7  & 64.1  & \textbf{80.4}    & 74.2        \\
breast color     & 42.8  & 61.1  & 77.4    & \textbf{90.4}        \\
belly color      & 56.6  & 65.7  & 81.3    & \textbf{93.3}        \\
bill length      & 60.3  & 55.8  & 84.3    & \textbf{95.5}    \\
\bottomrule
\end{tabular}
\caption{F1 scores for each body part. The issue-sensitive models are superior for all categories except `leg color'.}
\label{tab:body-part-aspect-per-image}
\end{table}

\begin{table}[tp]
\centering
\setlength{\tabcolsep}{8pt}
\begin{tabular}{lccc}
\toprule
&  Precision & Recall & $F_1$\\
\midrule
 $S_0$ & 10.5 &  21.1  & 15.5   \\
 $S_0$ Avg & \EDIT{12.1} & \EDIT{29.0} & \EDIT{17.0} \\
 $S_1$ & 11.2 &  21.7  & 14.8   \\
 $\IssueSpeaker$ & \textbf{18.7} &  42.5  & \textbf{25.9} \\
 $\IssueEntropySpeaker$ & 16.6 & \textbf{46.6} & 24.5  \\
\bottomrule
\end{tabular}
\caption{Issue alignment results.}
\label{tab:resolution-results}
\end{table}

\subsection{Human Evaluation} \label{sec:human:studies}

We conducted a human evaluation of our models primarily to assess their issue sensitivity, but also to validate the classifier we used in the previous automatic evaluations.

\paragraph{Study Design} 

We randomly sampled 110 images. For each, we have five conditions: our four RSA agents and the human captions from CUB. The items were arranged in a Latin Square design to ensure that no participant saw two captions for the same image and every participant saw an even distribution of conditions. We recruited 105 participants using Mechanical Turk. Each item was completed by exactly one participant. We received 1,365 responses in total. 

Participants were presented with a question text and a caption and asked to use the caption to select an answer to the question or indicate that the caption did not provide an answer. (No images were shown, of course, to ensure that only the caption was used.) For additional details, see Appendix~\ref{app:study-design}.

\begin{figure*}[htp]
\centering
\includegraphics[width=0.95\linewidth]{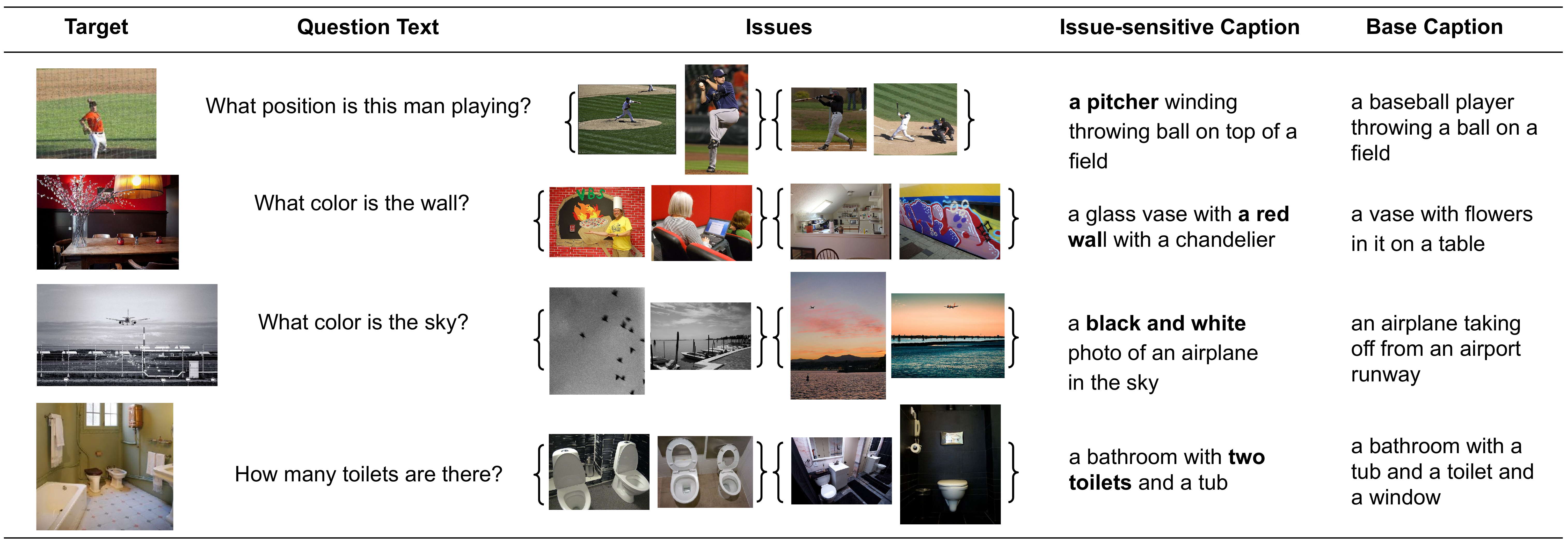}
    \caption{Captions for MS~COCO with issues determined by VQA~2.0. The left-hand cell in the `Issues' column contains the target image, and right-hand cell is the union of the distractor cells. The `Base Caption' texts are those produced by our $S_{0}$ model, and the `Issue-sensitive Caption' texts were produced by $\IssueEntropySpeaker$.}
\label{fig:mscoco-examples}
\end{figure*}

\paragraph{Issue Sensitivity}

\begin{table}[!tp]
\centering
\setlength{\tabcolsep}{8pt}
\begin{tabular}{l c c}
\toprule
 Caption Source &  Percentage & Size \\
\midrule
 $S_0$  & 20.9  & 273 \\
 $S_1$ &  24.5   & 273\\
 $\IssueSpeaker$ & 42.1 & 273\\
 $\IssueEntropySpeaker$ & \textbf{44.0} & 273  \\
 Human & 33.3 & 273 \\
\bottomrule
\end{tabular}
\caption{Percentage of captions that contain an answer to the question according to our human evaluation. By Fisher's exact test, both issue-sensitive models ($\IssueSpeaker$, $\IssueEntropySpeaker$) are different from issue-insensitive models ($S_0$, $S_1$) and humans ($p < 0.05$), but we do not have evidence for a difference between the issue-insensitive models.
}
\label{tab:human-studies}
\end{table}

\Tabref{tab:human-studies} shows the percentage of captions that participants were able to use to answer the questions posed. The pragmatic models are clearly superior. (The human captions are not upper-bounds, since they were not created relative to issues and so cannot vary by issue.)

\begin{table}[tp]
\centering
\setlength{\tabcolsep}{8pt}
\begin{tabular}{l cc}
\toprule
 Caption Source&  Accuracy & Size\\
\midrule
 $S_0$  & 94.5  & 273 \\
 $S_1$ &  90.1 & 273  \\
 $\IssueSpeaker$ & 92.7 & 273 \\
 $\IssueEntropySpeaker$ & 94.9 & 273  \\
 Human & 90.5 & 273 \\
\bottomrule
\end{tabular}
\caption{Classifier accuracy according to our human evaluation.}
\label{tab:classifier-fairness}
\end{table}

\paragraph{Classifier Fairness} We can also use our human evaluation to assess the fairness of the feature-in-text classifier (\secref{sec:cub:prelims}) that we used for our automatic evaluations. To do this, we say that the classifier is correct for an example $x$ if it agrees with the human response for $x$. \Tabref{tab:classifier-fairness} presents these results. Not only are accuracy values very high, but they are similar for $S_0$ and $\IssueEntropySpeaker$.

\section{MS~COCO and VQA 2.0}\label{sec:mscoco}

The annotations in the CUB dataset allow us to generate nuanced issues that are tightly connected to the content of the images. It is rare to have this level of detail in an image dataset, so it is important to show that our method is applicable to less controlled, broader coverage datasets as well. As a first step in this direction, we now show how to apply our method using the VQA~2.0 dataset \cite{goyal2017making}, which extends MS~COCO \cite{lin2014microsoft} with the question and answer annotations needed for VQA. While MS~COCO does have instance-level annotations, they are mostly general category labels, so the attribute-dependent method we used for CUB isn't effective here. However, VQA  offers a benefit: one can now control captions by generating issues from questions.

\paragraph{Dataset}

MS~COCO contains 328k images that are annotated with instance-level information. The images are mostly everyday objects and scenes.  A subset of them (204,721 examples) are annotated with whole image captions. \citet{antol2015vqa} built on this resource to create a VQA dataset, and \citet{goyal2017making} further extended that work to create VQA~2.0, which reduces certain linguistic biases that made aspects of the initial VQA task artificially easy. VQA 2.0 provides 1,105,904 question annotations for all the images from MS~COCO.

\paragraph{Constructing Partitions}

To generate issues, we rely on the ground-truth questions and answers in the VQA 2.0 dataset. Here, each image is already mapped to a list of questions and corresponding answers. Given an MS~COCO image and a VQA question, we identify all images associated with that question by exact string match and then partition these images into cells according to their ground-truth answers. Exactly the same procedure could be run using a trained VQA model rather than the ground-truth annotations in VQA 2.0.

\paragraph{Base Captioning System}

We use a pretrained state-of-the-art Transformer model with self-critical sequence training~\cite{rennie2017self}.
This has 6 Transformer layers with a 2048-dimensional hidden states, 512-dimensional input embeddings, and 8 attention heads at each layer. We use image features extracted by \citet{anderson2018bottom}. The model achieves a CIDEr score of 1.29 for the test split. We use beam search (with beam size 5) to generate our captions.

\paragraph{Example Captions}

We show some examples for MS~COCO in \figref{fig:mscoco-examples}. We chose these to highlight the potential of our model as well as remaining challenges. In datasets like this, the captioning model must reason about a large number of diverse issues, from objects and their attributes to more abstract concepts like types of food, sports positions, and relative distances (``How far can the man ride the bike?''; answer: ``Far''). Our model does key into some abstract issues (e.g.,  ``black and white photo'' in row~2 of  \figref{fig:mscoco-examples}), but more work needs to be done. \figref{fig:mscoco-examples} also suggests shortcomings concerning over-informativity (e.g., the mention of a tub in response to an issue about toilets).

 \section{Conclusion}\label{sec:conclusion}

We defined the task of Issue-Sensitive Image Captioning (\OurTask) and developed a Bayesian pragmatic model that allows us to address this task successfully using existing datasets and pretrained image captioning systems. We see two natural extensions of this approach that might be explored.

\EDIT{First, the method we proposed can be used as a method for assessing the quality of the underlying caption model. Using a dataset with issue annotations, if the model trained over the plain captions is more issue-sensitive, then it is better at decomposing the content of an image by its objects and abstract concepts.}

Second, one could extend our notion of issue-sensitivity to other domains. As we saw in \secref{sec:mscoco}, questions (as texts) naturally give rise to issues in our sense where the domain is sufficiently structured, so these ideas might find applicability in the context of question answering and other areas of controllable natural language generation.

\section*{Acknowledgement}

We thank the EMNLP referees for their constructive reviews. We also thank John Hewitt, Qi Peng, and Yuhao Zhang for feedback on an early draft of this paper. And special thanks to our Mechanical Turk participants for their crucial contributions.

\bibliography{anthology,emnlp2020}
\bibliographystyle{acl_natbib}

\appendix

\setcounter{figure}{0}
\renewcommand{\thefigure}{A\arabic{figure}}
\setcounter{table}{0}  \renewcommand{\thetable}{A\arabic{table}}
\appendix

\section{Incremental Pragmatic Reasoning}\label{app:incremental}

The normalization terms of $S_1$, $\IssueSpeaker$, and $\IssueEntropySpeaker$ all require a sum over all messages, rendering them intractable to compute. 

We now describe a variant of the $S_1$ which performs pragmatic reasoning incrementally. This method extends in an obvious fashion to $\IssueSpeaker$ and $\IssueEntropySpeaker$.

We begin by noting that a neural captioning model, at decoding time, generates a caption $\msg$ one segment at a time (depending on the architecture, this segment may be a word, word piece, or character). We write $\msg=(w_1 \ldots w_n)$, where $w_i$ is the $i$th segment.

Concretely, a trained neural image captioner can be specified as a distribution over the subsequent segment given the image and previous words, which we write as $S_0(w_{n+1}\given \state,[w_1\cdots w_n])$. This allows us to define incremental versions of $L_1$ and $S_1$, as follows:
\begin{multline}\label{eq:S1:inc}
L_{1}(\state\given w_{n+1},[w_1\ldots w_n]) \propto\\
  S_{0}(w_{n+1}\given\state,[w_1\ldots w_n])P(\state)
\end{multline}
\begin{multline}
U_1(\state,w_{n+1},[w_1,w_n]) =\\ \log
  L_{1}(\state\given w_{n+1},[w_1\ldots w_n])
\end{multline}
\begin{multline}
S_1(w_{n+1} \given \state,[w_1\ldots w_n])  
\propto\\ \exp(\alpha U_1(\state,w_{n+1},[w_1,w_n])-\mathit{cost}(w_{n+1}))
\end{multline}

Here, we define the cost as the negative log-likelihood of the $S_0$ producing $w_{n+1}$ given the image $i$ and previous segments $[w_1 \ldots w_n]$. We can then obtain a caption-level model, which we term $S_1^{\mathit{INC}}$ by contrast to the $S_1$ defined in (\ref{eq:S1-simple}):
\begin{align}
S_1^{\mathit{INC}}(\msg \given \state) =
\prod_{i=1}^{n} S_1(w_{i} \given [w_{1} \ldots w_{i-1}],\state)\label{SONESENTLP}
\end{align}
$S_1^{\mathit{INC}}(\msg\given\state)$ then serves as a tractable approximation of the caption-level $S_1(\msg\given\state)$, and the same approach is easily extended to $\IssueSpeaker$ and $\IssueEntropySpeaker$.

\section{CUB Attribute Annotation Example}\label{app:cub}

\begin{figure}[H]
\centering
\includegraphics[width=0.8\linewidth]{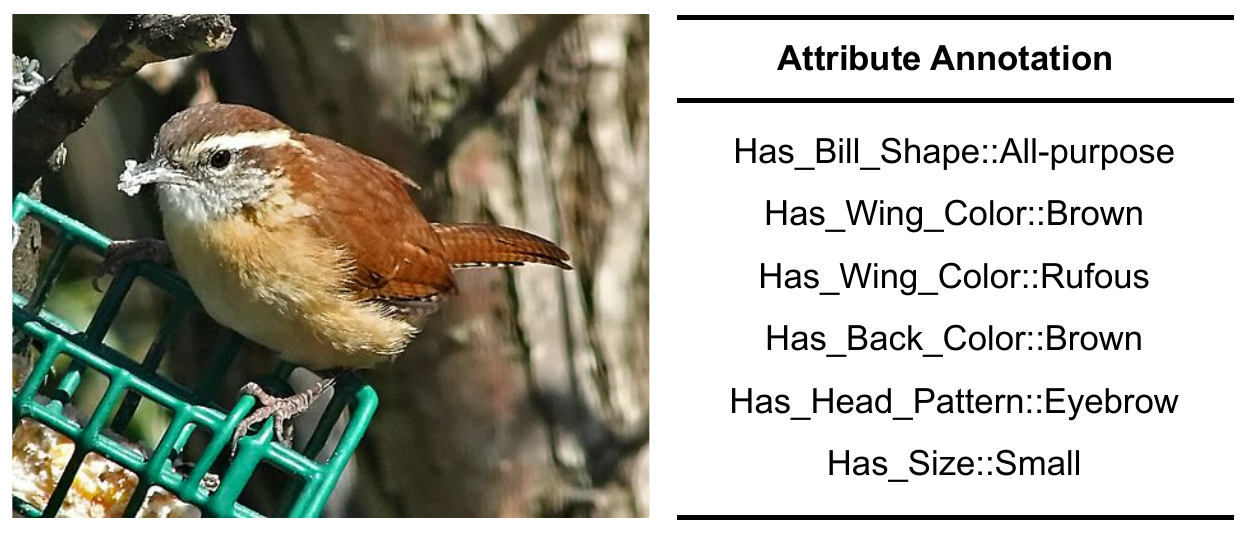}
\caption{A Carolina Wren from CUB. There can be multiple aspects per body part. Some general descriptors (e.g., \emph{size}) do not have fine-grained aspects.}
\label{fig:carolina-wren-attrs}
\end{figure}

\section{Human Study Design}\label{app:study-design}

Our study involved 110 randomly sampled images from CUB. For each, we have five conditions: original human caption, $S_0$ (base image caption model), $S_1$, $\IssueSpeaker$, and $\IssueEntropySpeaker$. The items were arranged in a Latin Square design to ensure that no participant saw two captions for the same image and every participant saw an even distribution of conditions. We recruited 105 participants using Mechanical Turk. Each participant completed 13 items, and each item was completed by exactly one participant. We received a total of 1,365 responses. No participants or responses were excluded from our analyses.

\begin{figure*}[tp]
\centering
\includegraphics[width=1\linewidth]{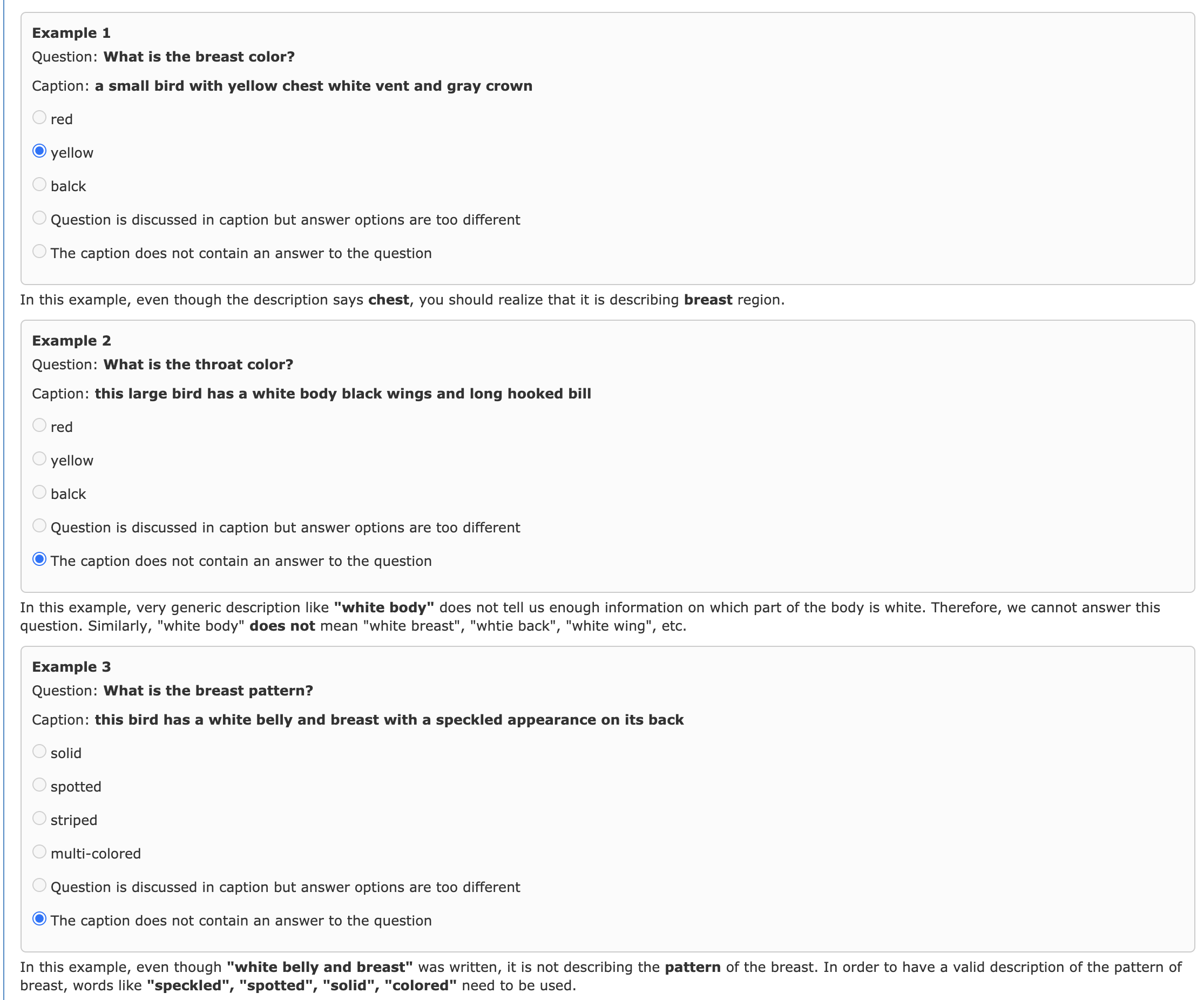}
    \caption{Example phase of the MTurk study}
\label{fig:mturk-screen-2}
\end{figure*}

In an instruction phase, participants were shown labeled images of birds to help with specialized terminology and concepts in the CUB domain. Before beginning the study, they were shown three examples of our task (\figref{fig:mturk-screen-2}). These examples were chosen to familiarize participants with the underlying semantics of the captions. For example, if caption gives an ambiguous generic description like ``the bird has a white body'', it does not provide enough information to answer questions about the color of specific body parts, even though it does mention a color. 

\begin{figure*}[tp]
\centering
\includegraphics[width=1\linewidth]{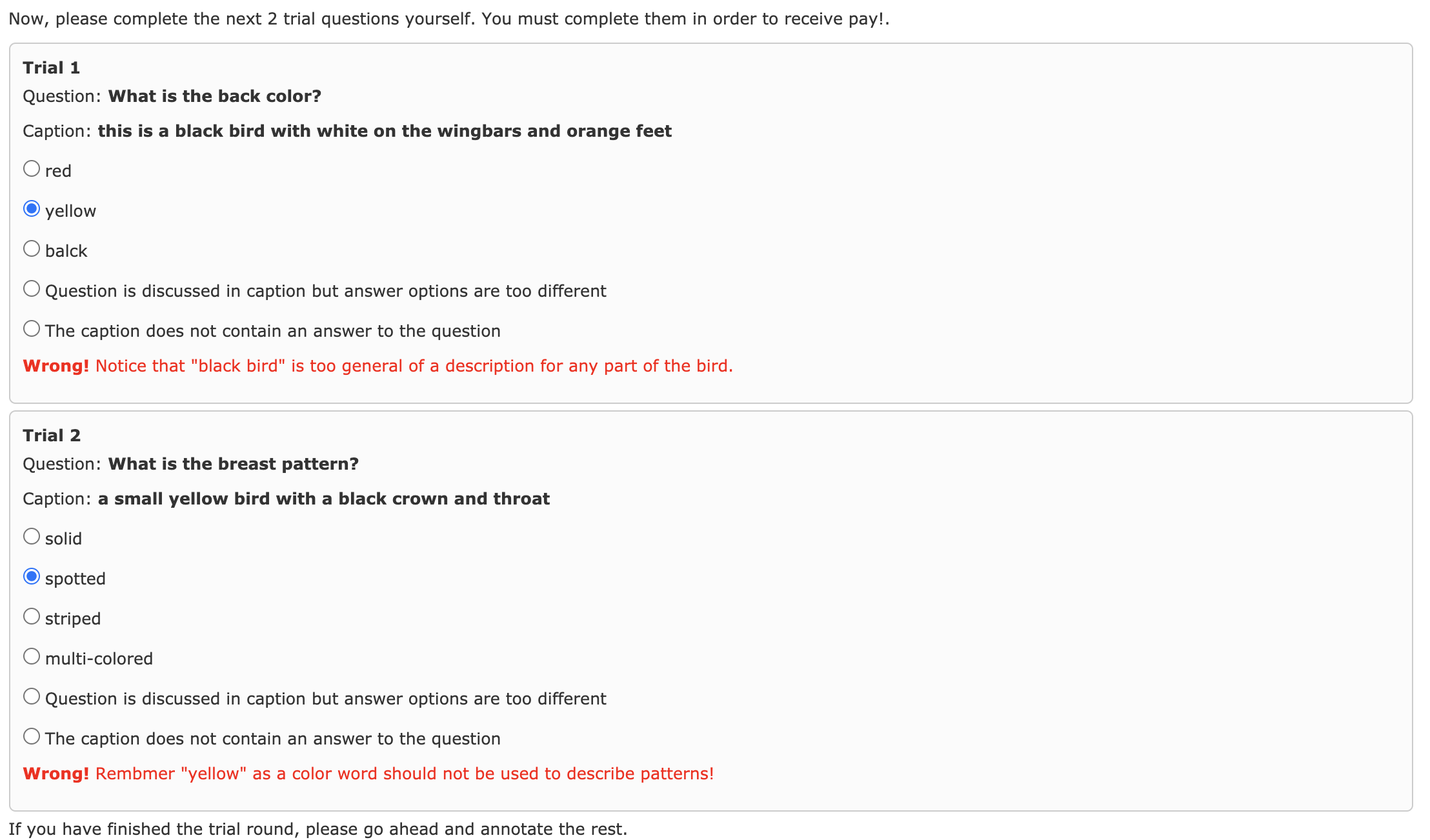}
    \caption{Trial phase of the MTurk study}
\label{fig:mturk-screen-3}
\end{figure*}

Following the example phase, we included a short trial phase of two example items, shown in \figref{fig:mturk-screen-3}. We required participants to complete this trial before they started the study itself. We provided feedback immediately (``Wrong'' or ``'Correct') after they made selections in this phase.

\begin{figure*}[!ht]
\centering
\includegraphics[width=0.7\linewidth]{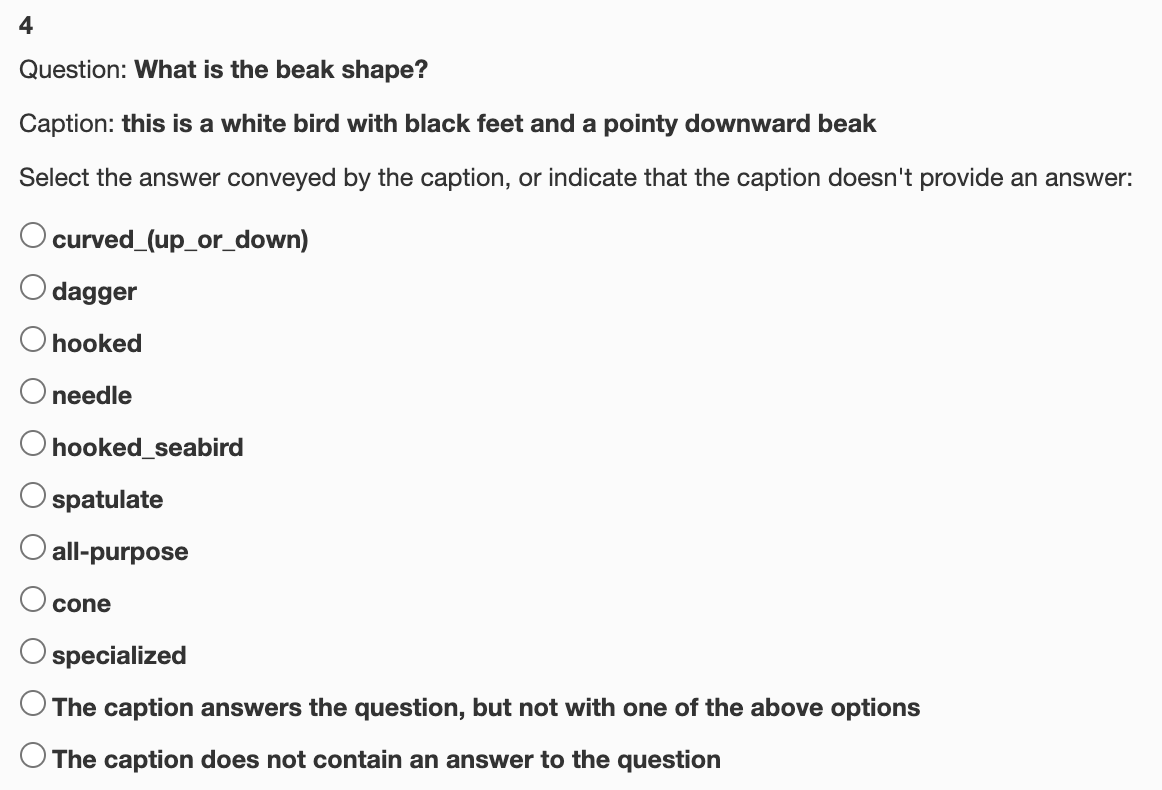}
    \caption{An annotation item from the MTurk study. In this example, the correct answer is ``curved (up or down)''.}
\label{fig:mturk-screen-5}
\end{figure*}

Finally, an example item is given in \figref{fig:mturk-screen-5}. Each annotation is structured in terms of a question (we rephrase issues in CUB as a question: \textit{has\_wing\_color} is rephrased as \textit{What is the wing color?}). Since the CUB attribute annotations consist of a \textit{property::value} structure, we take the values associated with the property as our answer options.

\section{Optimization Details}\label{app:optimization}

\paragraph{Computing infrastructure} Our experiment on CUB and MSCOCO is conducted on an NVIDIA TITAN X (Pascal) with 12196 MB graphic memory.

\paragraph{Computing time} Since we do not re-train our model, we report the inference time for our algorithm. Running our $\IssueEntropySpeaker$ on CUB test examples (5,794 images) takes about 40--50 minutes on a single GPU with specs listed above. For MS~COCO, it takes substantially longer (about 1 minute per image) due to the more complex Transformer base image captioning model.

\paragraph{Hyperparameters} We did not conduct a hyperparameter search. We manually set hyperparameters for our RSA-based models. The hyperparameters include rationality $\alpha$, entropy penalty $\beta$, and number of examples in a partition cell. The hyperparameters are chosen by small scale trial-and-error on validation data. We looked at the generated captions for 4 or 5 validation images of each model, and we decreased or increased our hyperparameters so that the generated captions for these images were coherent, grammatical, and issue-sensitive (when applicable). In \tabref{tab:hyperparam}, we report the hyperparameters we used for each RSA model.

\begin{table}[H]
\centering
\setlength{\tabcolsep}{8pt}
\begin{tabular}{l ccc}
\toprule
 Model &  Cell Size & $\alpha$ & $\beta$\\
\midrule
 $S_1$ &  40 & 3 & ---  \\
 $\IssueSpeaker$ & 40 & 10 & --- \\
 $\IssueEntropySpeaker$ & 40 & 10 & 0.4  \\
\bottomrule
\end{tabular}
\caption{Hyperparameter for each RSA model.}
\label{tab:hyperparam}
\end{table}

\paragraph{Validation performance}
We report the performance on the validation set in \figref{tab:attribute-coverage-val} and \figref{tab:resolution-results-val}, which contains 1,994 images. Even though the absolute numbers are slightly different from our result from the test set, the general trend still holds: issue-sensitive models significantly outperform issue-insensitive models.

\begin{table}[H]
\centering
\begin{tabular}{l c c c}
\toprule
   &  Precision & Recall & F1 \\
\midrule
$S_0$                  & \textbf{95.6} & 16.1 & 27.5  \\
$S_1$                  & 93.6 & 27.8 & 42.9  \\
$\IssueSpeaker$        & 94.8 & 41.6 & 57.8 \\
$\IssueEntropySpeaker$ & 94.4 & \textbf{43.7} & \textbf{59.7} \\ 
\bottomrule
\end{tabular}
\caption{Attribute coverage results for validation set.} \label{tab:attribute-coverage-val}
\end{table}

\begin{table}[H]
\centering
\setlength{\tabcolsep}{8pt}
\begin{tabular}{lccc}
\toprule
&  Precision & Recall & $F_1$\\
\midrule
 $S_0$ & 10.8 &  28.1  & 15.6   \\
 $S_1$ & 11.0 &  20.8  & 14.4   \\
 $\IssueSpeaker$ & \textbf{18.5} &  41.0  & \textbf{25.5} \\
 $\IssueEntropySpeaker$ & 14.8 & \textbf{42.4} & 22.0  \\
\bottomrule
\end{tabular}
\caption{Issue alignment results for validation set.}
\label{tab:resolution-results-val}
\end{table}

\section{More Examples in CUB}

We randomly sampled two test-set images (\figref{fig:cub-example-1} amd \figref{fig:cub-example-2}) to show qualitatively how well our issue-sensitive caption model does compared to other models. To increase readability in these figures, we gloss issues as question texts.

\begin{figure}[!t]
    \centering
    \footnotesize
    \begin{tabular}{|p{7.3cm}|}
\hline
\textbf{Image ID:} 121.Grasshopper Sparrow - Grasshopper Sparrow 0078 116052.jpg \\
\textbf{Human caption:} Small bird black tan and white feathers with a speck of yellow above beak. \\
\textbf{$S_0$ caption:} This bird has a white belly and breast with a speckled appearance on its back. \\
\hline
\textbf{What is the eye/eyering/eyebrow color?} \\
\textbf{$S_1$ caption:} A small tan and tan patterend bird with light belly. \\
\textbf{$\IssueSpeaker$ caption:} This is a littl brown bird with black stripes on the crown and superciliary. \\
\textbf{$\IssueEntropySpeaker$ caption:} This is a bird with a white belly and a black and white spotted back. \\
\hline
\textbf{What is the beak color?} \\
\textbf{$S_1$ caption:} This bird has a white belly with dark spots spots and a short pointy beak. \\
\textbf{$\IssueSpeaker$ caption:} This is a white and grey bird with an orange eyebrow and orange feet. \\
\textbf{$\IssueEntropySpeaker$ caption:} This is a small bird with a white belly and a brown back. \\
\hline
\textbf{What is the breast color?} \\
\textbf{$S_1$ caption:} A tan and tan sparrow connects to a white belly with tints of tan. \\
\textbf{$\IssueSpeaker$ caption:} A small bird with a white belly and throat and a spotted brown back and head. \\
\textbf{$\IssueEntropySpeaker$ caption:} This bird has a white belly and breast with a brown crown and short pointy bill. \\
\hline
\textbf{What is the belly color?} \\
\textbf{$S_1$ caption:} A small light brown and white bird with dark eyes and a short red-tipped bill. \\
\textbf{$\IssueSpeaker$ caption:} This bird has a white belly and breast with a speckled appearance elsewhere. \\
\textbf{$\IssueEntropySpeaker$ caption:} A small bird with a white belly and breast and a light brown crown and nape. \\
\hline
\textbf{What is the beak length?} \\
\textbf{$S_1$ caption:} A small round bird with multicolored tan and tan feathers. \\
\textbf{$\IssueSpeaker$ caption:} This is a brown and tan speckled bird with a small beak and long tail feathers. \\
\textbf{$\IssueEntropySpeaker$ caption:} This bird has a speckled belly and breast with a short pointy bill. \\
\hline
    \end{tabular}
    \caption{Issue-sensitive and Issue-insensitive captions for an image of a Grasshopper Sparrow. To increase readability in these figures, we gloss issues as question texts.}
    \label{fig:cub-example-1}
\end{figure}

\begin{figure}[!t]
    \centering
    \footnotesize
    \begin{tabular}{|p{7.3cm}|}
\hline
\textbf{Image ID:} 030.Fish Crow - Fish Crow 0073 25977.jpg \\
\textbf{Human caption:} A large all black bird with a fluffy throat and thick round beak. \\
\textbf{$S_0$ caption:} This bird is completely black with a long blunt bill. \\
\hline
\textbf{What is the crown color?} \\
\textbf{$S_1$ caption:} An all black crow black shiny black bill legs feet and body. \\
\textbf{$\IssueSpeaker$ caption:} This bird is all black with a long hooked bill and a think wingspan. \\
\textbf{$\IssueEntropySpeaker$ caption:} This is a mostly completely solid color bird with a black crown and a pointed bill. \\
\hline
\textbf{What is the beak color?} \\
\textbf{$S_1$ caption:} An all black crow black shiny black beak legs legs to shiny crow. \\
\textbf{$\IssueSpeaker$ caption:} An all black crow with strong thick downward downward curved black beak and black legs. \\
\textbf{$\IssueEntropySpeaker$ caption:} This is a small pointy bird with a medium sized beak and is mostly black with a short beak. \\
\hline
\textbf{What is the breast color?} \\
\textbf{$S_1$ caption:} An all black crow black bill legs feet and body. \\
\textbf{$\IssueSpeaker$ caption:} This stoutly black bird has strong legs and a long black beak. \\
\textbf{$\IssueEntropySpeaker$ caption:} This bird has a short black bill a white throat and a dark brown crown. \\
\hline
\textbf{What is the belly color?} \\
\textbf{$S_1$ caption:} An all jet black shiny black beak. \\
\textbf{$\IssueSpeaker$ caption:} A solid black crow with strong claws and a trinagular jet-black belly. \\
\textbf{$\IssueEntropySpeaker$ caption:} This bird has a short bill a white belly and a a black crown. \\
\hline
    \end{tabular}
    \caption{Issue-sensitive and Issue-insensitive captions for an image of a Fish Crow. To increase readability in these figures, we gloss issues as question texts.}
    \label{fig:cub-example-2}
\end{figure}

\end{document}